\documentclass[review]{elsarticle}
\usepackage{lineno,hyperref}
\modulolinenumbers[5]
\usepackage{amsthm,amssymb}
\usepackage{amsmath}
\usepackage{graphicx}
\usepackage{multirow}
\usepackage{arydshln}
\usepackage[]{algorithm2e}
\usepackage{subcaption}
\usepackage{multirow}
\usepackage{color}

\theoremstyle{plain} 

\theoremstyle{definition}

\theoremstyle{remark}

\usepackage{dutchcal}

\begin{document}
\begin{frontmatter}

\title{A Convex formulation for linear discriminant analysis}
\author{Sai Vijay Kumar Surineela \fnref{fn1}}
    \ead{surineela.2@wright.edu}

\author{Prathyusha Kanakamalla \fnref{fn2}}
    \ead{kanakamalla.2@wright.edu}

\author{Harigovind Harikumar \fnref{fn3}}
    \ead{harikumar.2@wright.edu}
    
\author{Tomojit Ghosh \corref{cor1} \fnref{fn4}}
    \ead{tomojit.ghosh@wright.edu}
\affiliation[1]{organization={Wright State University}, addressline={3640 Colonel Glenn Hwy}, postcode={45435}, city={Dayton}, state={Ohio}, country={USA}}

\cortext[cor1]{Corresponding Author}
\fntext[fn1]{Department of Computer Science and Engineering.}


\begin{abstract}
We present a supervised dimensionality reduction technique called Convex Linear Discriminant Analysis (ConvexLDA). The proposed model optimizes a multi-objective cost function by balancing two complementary terms. The first term pulls the samples of a class towards its centroid by minimizing a sample's distance from its class-centroid in low dimensional space. 
The second term pushes the classes far apart by maximizing their hyperellipsoid scattering volume via the logarithm of the determinant (\textit{log det}) of the outer product matrix formed by the low-dimensional class-centroids. Using the negative of the \textit{log det}, we pose the final cost as a minimization problem, which balances the two terms using a hyper-parameter $\lambda$. We demonstrate that the cost function is convex. Unlike Fisher LDA, the proposed method doesn't require to compute the inverse of a matrix, hence avoiding any ill-conditioned problem where data dimension is very high, e.g. RNA-seq data. ConvexLDA doesn't require pair-wise distance calculation, making it faster and more easily scalable. Moreover, the convex nature of the cost function ensures global optimality, enhancing the reliability of the learned embedding. Our experimental evaluation demonstrates that ConvexLDA outperforms several popular linear discriminant analysis (LDA)-based methods on a range of high-dimensional biological data, image data sets, etc.
\end{abstract}

\begin{keyword}
\texttt{ConvexLDA, Linear Discriminant Analysis, Convex Optimization, Linear Dimensionality Reduction, Supervised Dimensionality Reduction}
\end{keyword}

\end{frontmatter}


\section{Introduction}
\label{intro}

Dimensionality reduction (DR) plays a crucial role in pattern recognition tasks, enabling more effective data visualization, feature extraction, and improved downstream predictive performance \cite{ghosh2024linear,zhou2022supervised}. The recent surge in multi-source data collection has drastically increased data dimensionality, particularly in \textit{omics} analysis, where gene expression data from microarrays or next-generation sequencing can exceed 50,000 measurements \cite{tzec2025bioinformatics}.
High-dimensional datasets often contain noisy, redundant, missing, or irrelevant features, which can degrade the performance of pattern recognition tasks \cite{shen2022classification}. The acquisition of such high-dimensional datasets necessitates innovative techniques that can effectively handle large-scale data while remaining robust to noise \cite{tang2016visualizing}. DR is widely applied as an essential step to extract meaningful features enabling more effective data visualization, feature extraction, and improved downstream predictive performance \cite{ghosh2022supervised,migenda2024adaptive}.


With the advent of deep neural networks (DNNs) such as large language models (LLMs), convolutional neural networks (CNNs), and transformers, DR techniques may seem less prominent. However, despite the success of these complex architectures, linear dimensionality reduction remains a powerful and practical approach due to its interpretability, computational efficiency, and robustness in high-dimensional, low-sample-size (HDLSS) regimes \cite{shen2022classification}.

Deep learning models excel at learning hierarchical representations but pose significant challenges. They require large amounts of labeled data, extensive hyper-parameter tuning, and substantial computational resources. Additionally, these models often function as black boxes, offering little interpretability of their decision-making processes \cite{rudin2019stop}. In contrast, linear DR methods such as Linear Discriminant Analysis (LDA) and supervised Principal Component Analysis (SPCA) provide a more interpretable alternative that generalizes well even with limited data \cite{ghosh2024linear}.


Linear DR methods are particularly advantageous in small-data scenarios, where deep models tend to overfit \cite{zhang2019deep}. These linear techniques project data onto a low-dimensional subspace, preserving essential properties such as variance (as in PCA) \cite{jolliffe_1986} , class separability (as in LDA) \cite{fisher36lda} or maintaining pair-wise distances (as in MDS ) \cite{cox2000multidimensional}. Such projections enable visualization in two or three dimensions, offering intuitive insights into the structure of high-dimensional data. Most techniques in this category have a closed-form solution, which makes them attractive besides their simplicity. Furthermore, DR techniques improve the efficiency and accuracy of downstream pattern analysis tasks by eliminating noise, reducing redundancy, and enhancing class separability.

Despite its advantages, traditional LDA suffers from inherent limitations, including class-collapse in low-dimensional space \cite{hamsici2008bayes}, singularity of within-class scatter matrix in high-dimensional settings \cite{witten2011penalized}, and a lack of robustness to noise \cite{liu2024towards}. Additionally, LDA assumes that the optimal projection dimension is constrained to (number of classes -1), which may not always be suitable for complex datasets. In this work, we propose ConvexLDA, a convex formulation of LDA that overcomes these drawbacks while faithfully project classes by maintaining their grouping and separation. The contribution of our work is summarized below:


\begin{itemize}
    \item 
    Unlike Fisher’s LDA, which maximizes the ratio of between-class to within-class scatter, ConvexLDA minimizes a weighted sum of two components. The first component minimizes the distance between each sample and its class centroid in the reduced space, while the second component maximizes class separation by optimizing a weighted log-det function of the outer product of low-dimensional centroids.
    
    \item We demonstrate that the final cost is a convex optimization problem.
    
    \item  Unlike Fisher's LDA, our approach doesn't require to compute the inverse of within-class scatter ($S_w$) matrix. Hence it can avoids, situation arises from $S_w$ being singular when data dimension is high \cite{witten2011penalized}.

    \item The hyper-parameter $\lambda$ associated with the class-separation term can be tuned to push classes apart to avoid overlapping classes in low-dimension. This attribute of ConvexLDA mitigate the worst-case class separation problem of LDA \cite{wang2023worst}.

    \item The proposed ConvexLDA outperforms several state-of-the-art (SOTA) dimensionality reduction (DR) methods designed to address the limitations of Fisher’s LDA in pattern classification across various datasets, including real-world scRNA-Seq data. 
    
\end{itemize}

The remainder of this paper is structured as follows: Section \ref{lit_review} reviews existing LDA-based DR techniques and their limitations. Section \ref{methodology} presents our convex formulation of LDA, along with theoretical justifications. Section \ref{experiments} demonstrates the effectiveness of our approach through extensive benchmarking. Finally, Section \ref{conc_future_work} discusses key findings and future research directions.

\section{Related Work}
\label{lit_review}
Dimensionality reduction has a long history and remains an active area of research (see, e.g., \cite{nellas2023two} and references therein). Over the past decades, a variety of techniques have been developed, encompassing a broad spectrum of optimization frameworks and heuristic approaches. However, since our proposed method builds upon LDA, we focus our discussion on algorithms within this category.

Fisher's linear discriminant analysis (LDA) is historically one of the most widely used supervised dimensionality reduction techniques \cite{fisher36lda}. LDA reduces data-dimension by maximizing the ratio of the between-class scatter ($S_b$) to within-class scatter ($S_w$) that ultimately boils down to a generalized eigenvalue problem $S_b w = \lambda S_w w$. The problem can be translated into eigen decomposition in the form $({S_w}^{-1}S_b) w = \lambda w$. LDA creates the mapping for a $C$ class data using a $C-1$-dimensional sub-space. One issue arises in high dimension low sample size (HDLSS) cases when the $S_w$ becomes singular \cite{witten2011penalized}. Several approaches were taken to overcome the singularity issue. In direct LDA (DLDA) \cite{yu2001direct}, the authors suggested to project the original data to the range of the $S_b$ (thus discard the null space of $S_b$ which does not contain any discriminative information) followed by a transformation using a regulating matrices to find direction that minimizes with-in class scatter. Note, Gao et al. \cite{gao2006direct} argued that DLDA is not quite equivalent to LDA. In regularized LDA (RLDA) \cite{dai2007face,zheng2009perturbation}, a small perturbation is added to the $S_w$ to avoid the ill-condition. Belhumeur et al. \cite{belhumeur1997eigenfaces} proposed to project the data onto low-dimensional PCA subspace to make the $S_w$ full rank, followed by applying LDA. This approach is referred to PCA + LDA and a theoretical justification is provided in \cite{yang2003can}. In null space LDA (NLDA) \cite{chen2000new}, data is projected on to the null space of $S_w$ and then the maximization of $S_b$ is performed to get the final projection direction. NLDA tends to overfit the training data; however, its generalization performance can be improved using Representative Null Space LDA (RNLDA) \cite{he2021representative}.

Apart from the singularity issue, LDA has another problem where the model overlaps classes if the embedding dimension is less than the number of classes \cite{hamsici2008bayes}. This problem is known as \textit{worst-case class separation} problem. There have been several attempt to alleviate the issue. Zhang et al. proposed worst-case linear discriminant analysis (WLDA) \cite{zhang2010worst} which maximize the minimum between-class scatter and minimize the maximum within-class scatter. Although sounds a reasonable idea, optimizing the two extreme class scatter may be hard to achieve in real-world data sets. Contrary to WLDA, Yang et al., introduced a probabilistic model DR-MSP \cite{yang2019nonparametric} which maximizes the sum of probability among all class pairs instead of focusing on worst class pairs. DR-MSP failed to separate the worst-class pairs \cite{wang2023worst}. Many attempts were made to solve the issue by assigning weight while learning the discriminatory subspace. Pairwise Accuracy Criterion (aPAC) \cite{loog2001multiclass}, harmonic mean subspace selection (HMSS) \cite{bian2008harmonic}, geometric mean subspace selection (GMSS) \cite{tao2008geometric} falls in this category. These models assumes homoscedastic Gaussian assumption, which doesn't always hold in real-world data. Su et al. proposed (HMMDA) \cite{su2018heteroscedastic} using Chernoff distance instead of Euclidean to measure the separability of all class pairs and avoiding the homoscedastic Gaussian assumption of HMSS and GMSS. On the similar line of research, Zhu et al. proposed neighborhood LDA (nLDA) \cite{zhu2022neighborhood} to handle multi-modal classes. Liu et al. proposed a ratio trace formulation (WDA-eig) \cite{liu2020ratio} using the Wassterstein distance. A lot of research has been done from the point of view of  Max-Min Distance Analysis (MMDA) \cite{shao2017regularized,yu2011distance,zhang2010worst}. Bian and Tao proposed to maximize the distance of the worst class-pair using semidefinite programming (SDP) \cite{bian2010max}. As SDPs are computationally expensive, Li et al. \cite{li2015worst} proposed scalable optimization with a quasi-Newton technique and eigen-decomposition which is faster than standard interior-point SDP solvers. MMDA-based models can solve the worst-case class separation problem , but as pointed out by Wang et al. \cite{wang2023worst}, these models fail to control the class overlap issue completely when the with in-class scatter is large.  In their recent work, Worst-case Discriminative Dimension Reduction (WDDR) \cite{wang2023worst}, the authors proposed to optimize a ratio criteria, called Max-Min Ratio Analysis (MMRA) instead of MMDA. WDDR computes all the ratios of between-class scatter to joint within-class scatter of each class pairs and then maximizes the minimum quantity.

Many researchers have claimed that LDA is not robust in the presence of outlier due to its reliance on $\ell_2$-norm and proposed several variation using $\ell_1$-norm. Examples of this line of work can be found in \cite{lu2016l1,liu2024towards}.    

\section{Convex Formulation of Linear Discriminant Analysis (ConvexLDA)}
\label{methodology}

Let $X \in \mathbb{R}^{d \times n}$ is the data matrix with $n$ samples where each sample $x_i$ is a $d$-dimensional vector ($x_i \in \mathbb{R}^d$). Assume $X$ has $M$ number of classes $\{C_j\}^M_{ j = 1}$
where the 
index of class $C_j$ is denoted by $I_j$.  We define centroid of each class as
\begin{equation}
c_j=\frac{1}{|C_j|}\sum_{i \in I_j} x_i
\end{equation}
where $|C_j|$ is the cardinality of  class $C_j$.  Let's define a matrix $\tilde{C} \in \mathbb{R}^{d \times n}$ which contains the corresponding $c_j$s for each sample $x_i$. Note $\tilde{C}$ will have non-unique entries. Let's have another matrix $\hat{C}$ which only contains the unique centroids $c_j$s of each class. More precisely, each column of $\hat{C}$ is a centroid $c_j$. For example, consider the data set $X=\{x_1,x_2,x_3,x_4,x_5\}$ which has two classes, say, $C_1,C_2$ where $I_1=\{1,3,5\} \;; I_2=\{2,4\}$ and $c_1,c_2$ are the corresponding centroids. In this case $\tilde C = \{c_1,c_2,c_1,c_2,c_1\}$ and $\hat C = \{c_1,c_2\}$.
Under this set up we seek a transformation matrix $A \in \mathbb{R}^{d \times p}$ to achieve the following two goals in the reduced space:\\
    1. each point should be mapped approximately to its class centroid\\
    2. the centroids should be maximally scattered

To achieve the first goal we minimize the quantity $\|A^T(\tilde{C}-X)\|^2_F$ where 
$\| .\|_F$ denotes the Frobenius norm. Minimizing this quantity will take a sample $x_i \in C_j$ close to its corresponding class centroid $c_j$ in reduced space. 
It is useful to use the fact $\|A^T(\tilde{C}-X)\|^2_F = 
Tr[(\tilde{C}-X)(\tilde C -X)^T AA^T]$. To achieve the second goal, we maximize $\log \det(A^T {\hat C}(A^T \hat C)^T + \gamma I)$ where $\gamma$ is small positive number 
(in our experiments we keep this value very small). 
The quantity $\det(A^T {\hat C}(A^T \hat C)^T)$ gives the square of the scattering volume of the hyper-ellipsoidal formed by the low dimensional centroids \cite{duda2006pattern}.
Maximizing this volume will scatter the centroids maximally in low dimensions. 
Taken together, we are proposing the following minimization problem over the transformation matrix $A$:
\begin{equation}
\begin{aligned}
\underset {A} {minimize}\;\;Tr[(\tilde{C}-X)(\tilde C -X)^T AA^T] - \lambda\log \det (A^T {\hat C}{\hat C}^T A + \gamma I) \\
\end{aligned}
\label{equation:CCR_objective}
\end{equation}
where $\lambda$ is a positive quantity which balances the two terms. The crucial property of this multi-objective optimization is that it's a convex function of the matrix $A$.

\subsection{Proof of Convexity}
We write the original objective as a sum to two parts, $\mathcal{L} = \mathcal{L}_1 \; + \; \lambda \mathcal{L}_2 $
where
$$\mathcal{L}_1 = Tr[(\tilde{C}-X)(\tilde C -X)^T AA^T]$$
$$\mathcal{L}_2= - \log \det (A^T {\hat C}(A^T \hat C)^T + \gamma I)
$$
First, observe that the domain, i.e., the set of 
$d \times p$ matrices, is a convex set.
Now we will show that $\mathcal{L}_1$ is convex. Differentiating $\mathcal{L}_1$ w.r.t. $A$ gives
\begin{equation}
\begin{aligned}
 \Delta \mathcal{L}_1 = 2(\tilde C -X)(\tilde C-X)^T A
\end{aligned}
\label{derivative_L1}
\end{equation}
Differentiating again
\begin{equation}
\begin{aligned}
 \Delta^2 \mathcal{L}_1 = 2(\tilde C -X)(\tilde C-X)^T
\end{aligned}
\end{equation}
Observe that the Hessian of $\mathcal{L}_1$ is the sample co-variance matrix therefore it's positive semidefinite. Hence by the second-order condition of convexity \cite{10.5555/993483}, $\mathcal{L}_1$ is a convex function.

Now we show $\mathcal{L}_2= - \log \det (A^T {\hat C}{\hat C}^T A + \gamma I )$ is convex. Observe that the  matrix $M = (A^T {\hat C}{\hat C}^T A + \gamma I )$ is a positive definite matrix when $\gamma > 0$.  
Consider the value of $\mathcal{L}_2$ on an arbitrary 
line segment given by $M = P + t Q$ where $P$ and $Q$ are positive definite matrices. Note that the domain of positive definite matrices is also convex.  Now define a function $g$ as 
\begin{equation}
\begin{aligned}
g(t) = - \log \det(P + t Q)
\end{aligned}
\label{tmpEq15}
\end{equation}
As $P$ is positive definite, we can write $P=P^{1/2}P^{1/2}$ where $P^{1/2}$ is also positive definite. Hence we can write,
\begin{equation}
\begin{aligned}
g(t) = - \log \det(P^{1/2}(I + t P^{-1/2} Q P^{-1/2})P^{1/2})
\end{aligned}
\label{tmpEq20}
\end{equation}
Using the fact $\det (AB)=\det A \det B$ and the properties
of logarithms we conclude
%
\begin{equation}
\begin{aligned}
g(t) = - \log \det P - \log \det(I + t P^{-1/2} Q P^{-1/2})
\end{aligned}
\label{tmpEq22}
\end{equation}
The $\det (I + tP^{-1/2} QP^{-1/2})$ is the product of the eigenvalues of $(I + tP^{-1/2} QP^{-1/2})$. Let $\sigma_1,\sigma_2,...,\sigma_n$ are the eigenvalues of $tP^{-1/2} QP^{-1/2}$. Hence,
\begin{equation}
\begin{aligned}
g(t) = - \log \det P - \log \prod^{n}_{i=1} (1+t\sigma_i)
\end{aligned}
\label{tmpEq23}
\end{equation}

\begin{equation}
\begin{aligned}
g(t) = - \log \det P - \sum^{n}_{i=1} \log(1+t\sigma_i)
\end{aligned}
\label{tmpEq24}
\end{equation}
Differentiating w.r.t. $t$ twice gives
$$\Delta^2 g(t) = \sum^{n}_{i=1} \frac{{\sigma_i}^2}{{1+t\sigma_i}^2} \ge 0$$
Since $g$ is convex we conclude $\mathcal {L}_2$ is convex. So the cost is a summation of two convex functions over a convex set, hence it's also convex function\footnote{The sum of two convex functions is also convex~\cite{10.5555/993483}}.

\subsection{Calculation of Gradient}
The gradient of the cost function Equation \ref{equation:CCR_objective} is given below:
\begin{equation}
\frac{\partial \mathcal{L}}{\partial A} = 2(\tilde{C} -X)(\tilde{C} -X)^TA - 2\lambda({\hat C}{\hat C}^T A)(A^T {\hat C}{\hat C}^T A+\gamma I)^{-1}
\label{equation:gradient_CCR}
\end{equation}
The derivation is left in supplementary material.

\section{Analysis}

\subsection{Experiment to analyze the effect of \texorpdfstring{$\lambda$}{Lg}}
\label{effect_of_lambda}
\begin{table}[!ht]
	\centering
	\begin{tabular} {|c|c|c|c|c|}	
		\hline 	
		\multicolumn{1}{|c|} {Hausdorff Dist.} & \multicolumn{1}{c|} {$\lambda = 0.1$} &  \multicolumn{1}{c|} {$\lambda = 1.0$}  & \multicolumn{1}{c|} {$\lambda = 10.0$} & \multicolumn{1}{c|} {$\lambda = 100.0$} \\
		\hline
		{Digit 0 vs Digit 1} & $10.9114$ & $13.4852$ & $17.9494$ & $44.3374$ \\
		\hline
		{Digit 1 vs Digit 0} & $0.5410$ & $1.6782$ & $5.2770$ & $16.6911$ \\
		\hline
		{Digit 0 vs Digit 2} & $9.6558$ & $11.7688$ & $11.9953$ & $28.9347$ \\
		\hline
		{Digit 2 vs Digit 0} & $343.7246$ & $473.9643$ & $475.3801$ & $498.0760$ \\
		\hline
		{Digit 1 vs Digit 2} & $0.4137$ & $1.3104$ & $4.1407$ & $13.0951$ \\
		\hline
		{Digit 2 vs Digit 1} & $353.1745$ & $485.9086$ & $483.7695$ & $503.1173$ \\
		\hline
    \end{tabular}
        \caption{The effect of $\lambda$ on a subset of MNIST.  The 3-dimensional data is represented in 2D by ConvexLDA. Hausdorff distances among the class pairs are measured for each value of $\lambda$.}
        \label{table:MNIST3class_hausdorffDist}
	
\end{table}

ConvexLDA requires the tuning of the hyper parameter $\lambda$ which is used to control the separation among the classes. A bigger value will make the classes more separated compared to a smaller one. To demonstrate this we conducted experiment with the following values of $\lambda$: $0.1,1.0,10.0,100.0$ on a subset of MNIST. We took all the samples from digit class 0, 1 and 2 and trained the model to get a two-dimensional representation of each sample. Then we calculated the Hausdorff distances of each pair of classes and present the result in Table \ref{table:MNIST3class_hausdorffDist}. It's clear that as the value of $\lambda$ increases the Hausdorff distances also increases thus increasing the class separation.

\subsection{Trade-off of the Optimization}
The experiments in Section \ref{effect_of_lambda} demonstrates that the class separation increases with the increase of $\lambda$. But increasing $\lambda$ will also increase the scatter of each classes. The explanation lies in the objective function. The term $\mathcal{L}_1$  brings the samples of a class near to its centroid while $\mathcal{L}_2$ serves to increase the separation among the classes. As we increase $\lambda$ the gradient of the second part of the objective function will start to dominate and the model will focus more on class-separation. As a result the scatter of each class will start to increase. To examine this, we ran an experiment using the same MNIST subset used in Section \ref{effect_of_lambda}. To objectively quantify the class-scatter and class-separation, we define two measures, class scatter (CS) and mean set distance (MSD) as below:
\begin{equation}
\begin{aligned}
CS(C_j) = \frac{1}{|C_j|} \sum_{x_i \in C_j} dist(c_j,x_i) \hspace{1.75 cm}\\
MSD(C_i,C_j) = \frac{1}{|C_i|} \sum_{x_k \in C_i} \frac{1}{|C_j|} \sum_{x_l \in C_j} dist(x_k,x_l)
\end{aligned}
\label{individual_classscatter}
\end{equation}
Note, MSD is calculated on a class pair as opposed to CS. Table \ref{table:MNIST3class_scatter} presents the CS values for each digit class over different values of $\lambda$.  Higher values of $\lambda$ increase the scatter. We also observe in Table \ref{table:MNIST3class_separation} that increasing $\lambda$ also increases the pair-wise class separation.

\begin{table}[!ht]
	\centering
	\begin{tabular} {|c|c|c|c|c|}	
		\hline 	
		{CS} & {$\lambda = 0.1$} &  {$\lambda = 1.0$}  & {$\lambda = 10.0$} & {$\lambda = 100.0$}\\
		\hline
		{Digit 0} & $2.8746$ & $4.0491$ & $6.6547$ & $11.7280$ \\
		\hline
		{Digit 1} & $5.0957$ & $9.0666$ & $16.1205$ & $28.6672$ \\
		\hline
		{Digit 2} & $13.1986$ & $15.5978$ & $18.2859$ & $27.1962$ \\
		\hline
	\end{tabular}
    \caption{The effect of $\lambda$ on class-scatter on a subset of MNIST digits. The $784$-dimensional images are projected in two-dimensional space to compute the CS scores.}
	\label{table:MNIST3class_scatter}
\end{table}

\begin{table}[!ht]
	\centering
	\begin{tabular} {|c|c|c|c|c|}	
		\hline 	
		{MSD} & {$\lambda = 0.1$} &  {$\lambda = 1.0$}  & {$\lambda = 10.0$} & {$\lambda = 100.0$} \\
		\hline
		{Digit 0 vs Digit 1} & $1.2339$ & $2.1895$ & $3.8876$ & $6.9136$ \\
		\hline
		{Digit 0 vs Digit 2} & $1.1300$ & $1.9907$ & $3.5181$ & $6.2386$ \\
		\hline
		{Digit 1 vs Digit 2} & $1.0188$ & $1.7988$ & $3.1827$ & $5.6430$ \\
		\hline
	\end{tabular}	
    \caption{The effect of $\lambda$ on class-separation on a subset of MNIST digits. First, the $784$ dimensional data is represented in 2D by our model. After that the CR and MSD are measured on 2D space.}
	\label{table:MNIST3class_separation}
\end{table}

\begin{figure}[!ht]
\hspace{-1.5cm}
	\includegraphics[width=14.0cm,height=9.0cm]{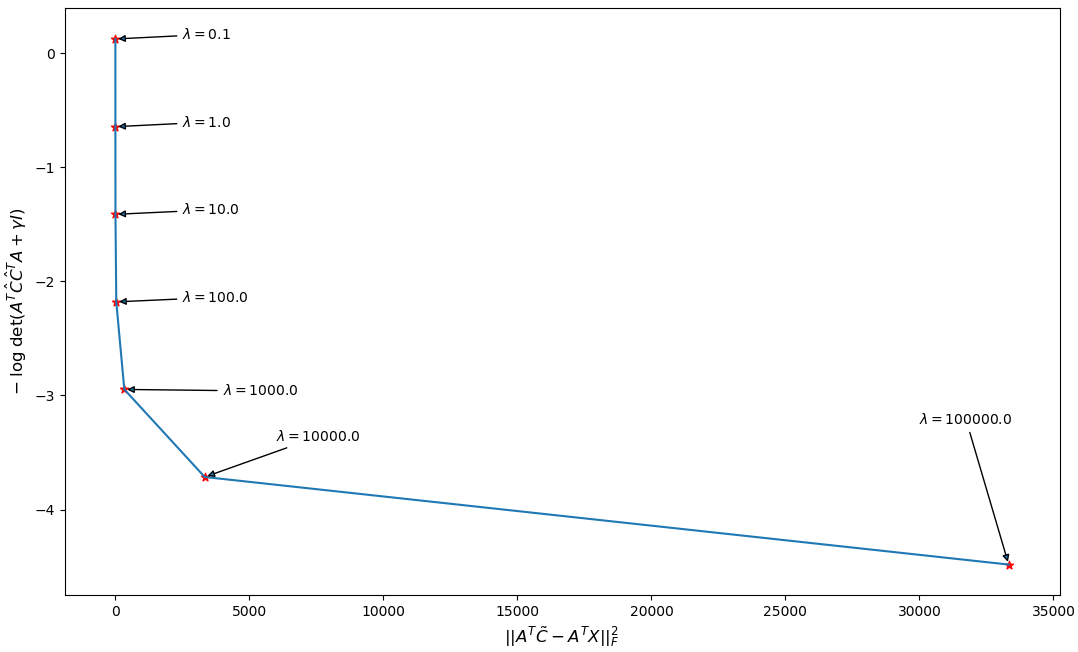}
	\caption{Plot of $\mathcal{L}_1$ and $\mathcal{L}_2$ across $\lambda$.}
	\label{fig:frontier_plot}
\end{figure}

We also plotted the sub-costs $\mathcal{L}_1$ and $\mathcal{L}_2$ for different values of $\lambda$ in Figure \ref{fig:frontier_plot}. Higher values of $\lambda$, increases $\mathcal{L}_1$, indicating that samples become less tightly clustered—an undesirable effect. Conversely, $\mathcal{L}_2$ decreases with increasing $\lambda$, signifying improved class separation, as expected. This highlights the fundamental trade-off in the model: $\lambda$ controls the balance between maintaining compact class clusters and enhancing inter-class separation. For prediction tasks, an optimal $\lambda$ can be selected by minimizing generalization error, such as $k$-NN classification error in the embedding space, on a validation set.

\subsection{Analysis of class collapse between Fisher LDA and ConvexLDA}
\begin{figure}[!ht]
	\includegraphics[width=12.0cm,height=10.0cm]{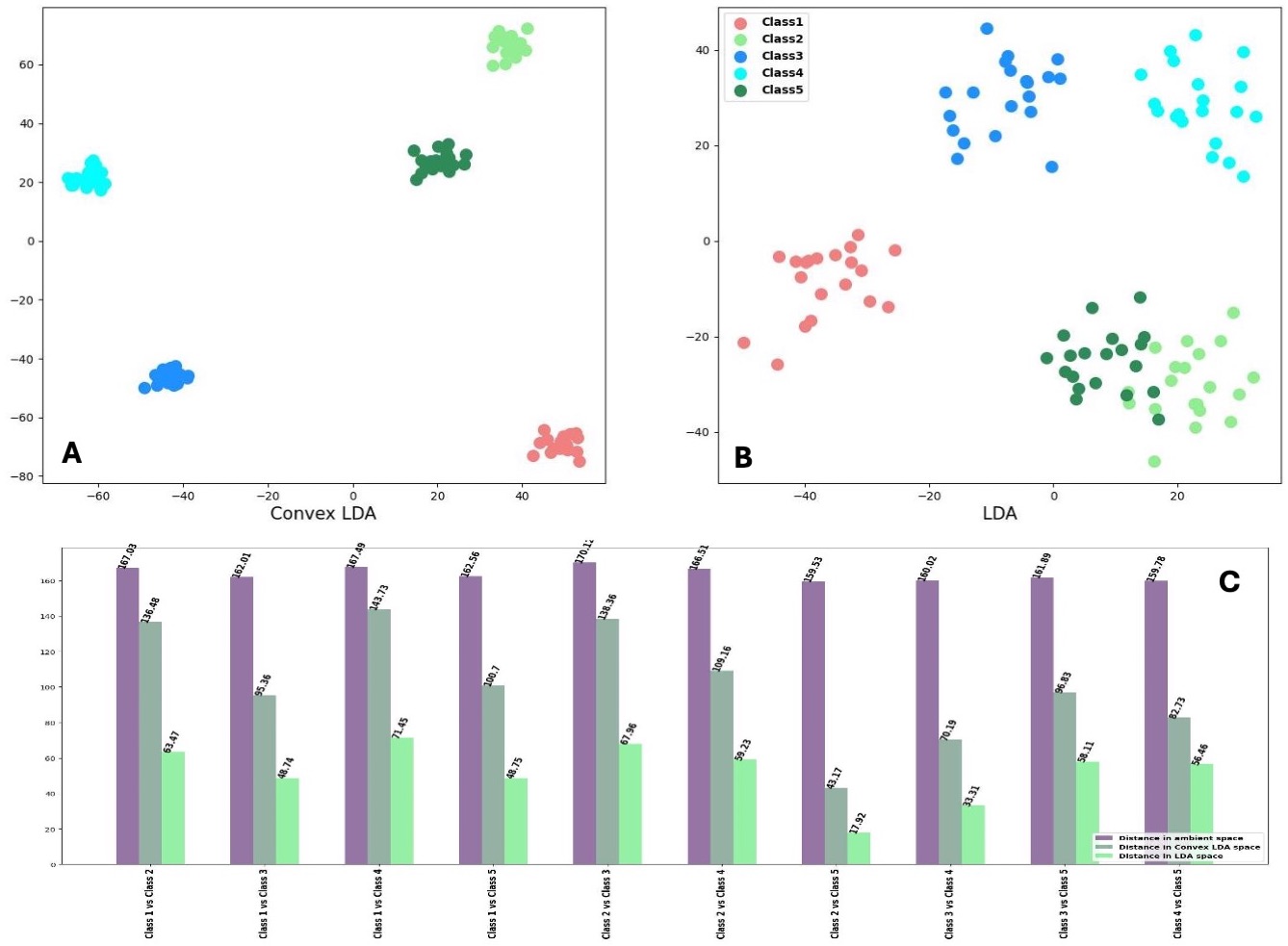}
	\caption{Visualization of synthetic samples using Fisher's LDA (panel 'B') and Convex LDA (panel 'A') in two-dimensional embedding space. Panel 'C' presents the distance among each class pairs in original, LDA, Convex-LDA space. ConvexLDA is run setting $\lambda$ to 1.}
	\label{fig:LDA_vs_ConvexLDA1}
\end{figure}

Class collapse or worst-class separation problem is a known issue of Fisher LDA and it happens when the projection dimension is less than the number of classes. This section provides empirical proof of how ConvexLDA mitigates the issue. To demonstrate it, we used a toy data set sampled from an isotropic Gaussian distribution with five classes where each sample belongs to $\mathbb{R}^{100}$-dimensional space. We randomly sampled a total of 100 points where the mean of each class is randomly selected and the standard deviation is set to 20. In Figure \ref{fig:LDA_vs_ConvexLDA1}, we show the projection of the 
samples in the two-dimensional space using the two methods along with the class-pair distances. Visually, it's clear that the samples of each class is more scattered in LDA space compared to our proposed technique. Notice, LDA mixes the samples of classes 2 and 5 where ConvexLDA separates them with tighter cluster. It's noteworthy, that mixing of samples by LDA when the embedding dimension is lower than the number of classes is reported in literature, see \cite{hamsici2008bayes}, and we have encountered the same issue. In contrast, our proposed approach preserves the class-pair distance better, as depicted in panel 'C'. 
\begin{figure}[!ht]
	\includegraphics[width=12.0cm,height=8.0cm]{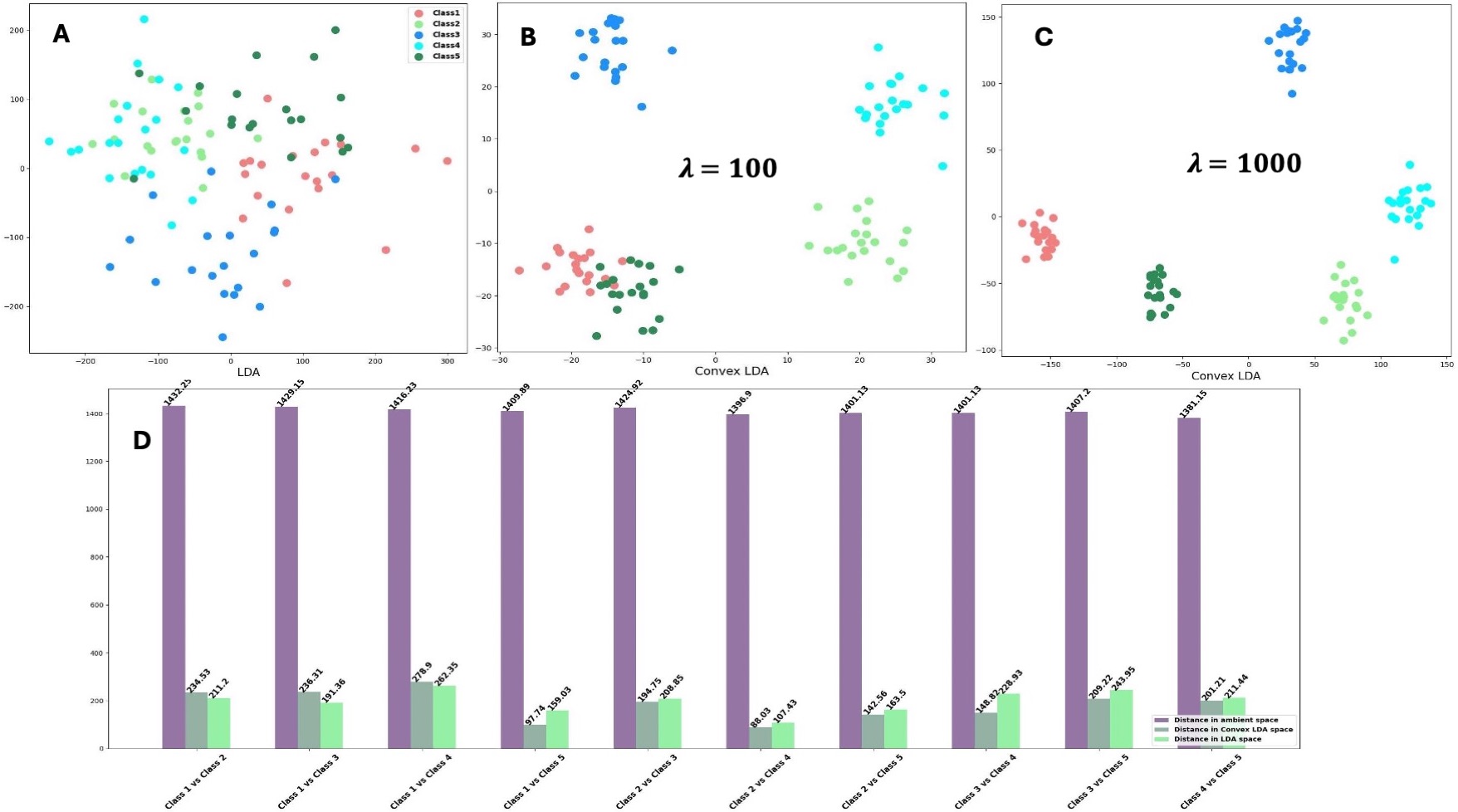}
	\caption{Visualization of toy data set in LDA and ConvexLDA space. In this experiment, we ran ConvexLDA with two different values of $\lambda$.}
	\label{fig:LDA_vs_ConvexLDA2}
\end{figure}

In the next experiment, we used slightly modified toy data where we increased the standard deviation of each class to 100. Again, we sampled 100 samples in total and ran LDA and ConvexLDA to project the data in two-dimensional space. Figure \ref{fig:LDA_vs_ConvexLDA2} shows the results in four panels. LDA didn't separate the classes at all, making one cluster where all the samples got mixed. ConvexLDA with $\lambda$=100 separate the classes excepts for classes 1 and 5. When we increased the $\lambda$ to 1000, we see a clear separation of all the five classes. In panel D, we present the class pair distances in original space, LDA and ConvexLDA ($\lambda$=1000). In some cases LDA has higher class-pair distances, but its inability to minimize the within-class variance makes the projection less informative. In these two toy examples, we demonstrate that our proposed model minimizes the within-class scatter better than LDA and at the same time it is also capable of finding a linear transformation that separates the classes to avoid mixing of samples. It is also noteworthy, that the choice of $\lambda$ plays an important role in class separation.

\subsection{Hyper-parameter tuning:}
Our model has two hyperparameters, the embedding dimension ($p$), and $\lambda$.
\begin{figure}[!ht]
	\includegraphics[width=12.0cm,height=2.25cm]{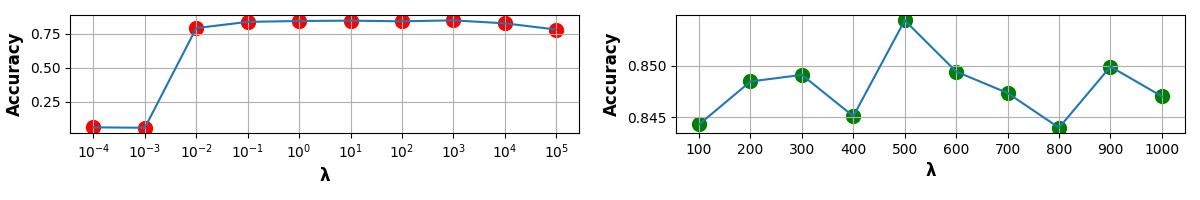}
	\caption{Optimal value of $\lambda$ for ISOLET data for embedding dimension 10.}
	\label{fig:ISOLET_optimal_lambda}
\end{figure}

\begin{figure}[!ht]
	\includegraphics[width=12.0cm,height=2.25cm]{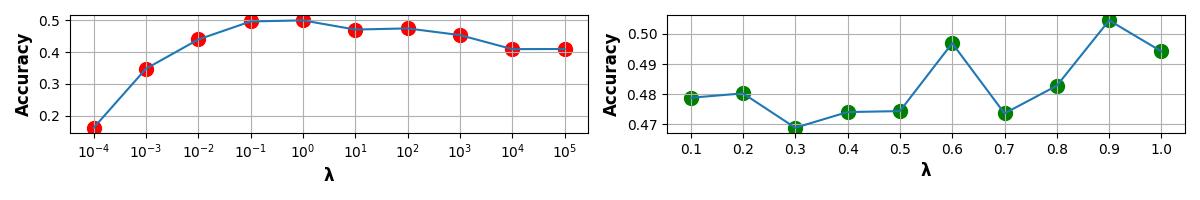}
	\caption{Optimal value of $\lambda$ for MNIST data for embedding dimension 2.}
	\label{fig:MNIST_optimal_lambda}
\end{figure}
We ran a five-fold cross-validation on the training partition to find the optimal values. In Figures \ref{fig:ISOLET_optimal_lambda} and \ref{fig:MNIST_optimal_lambda} we present two examples on ISOLET and MNIST for embedding dimensions $10$ and $2$ respectively. In both cases, we started the search for $\lambda$ on a logarithmic scale. For ISOLET, we see that 5-fold cross-validation accuracy started to saturate after $\lambda$ reaches to $1$, but there is a slight increase in accuracy from $10^2$ to $10^3$. A search between these two values in a linear scale gives us the optimal value of $500$. 
Similarly, for MNIST, the accuracy remains the same for $0.1$ and $1$ and a search using a linear interval of $0.1$ returns $0.9$ as the optimal value.

\section{Benchmarking Experiments and Results}
\label{experiments}
This section provides the details of the classification and visualization experiments and the results.
To objectively compare our proposed model, we benchmark it with LDA and several of its variants. The benchmarking is done on low-dimensional as well as on the optimal embedding dimension of each model. We used publicly available datasets which are frequently used to compare DR techniques in the literature. The description of these datasets are kept in Table \ref{table:dataDescription}. 
\begin{table}[!ht]	
	\centering
	\begin{tabular} {|c|c|c|c|c|c|}	
		\hline	
		Dataset & \#Features & \#Classes & \#Samples & Domain \\
		\hline
        USPS & 256 & 10 & 9298 &  \\
        MNIST & 784 & 10 & 60000 &  \\
        Olivetti & 4096 & 40 & 400 & \multirow{2}{*}{Image} \\
        Yale Face & 1024 & 15 & 165 &   \\
        COIL20 & 1024 & 20 & 1440 &  \\
        ORL & 1024 & 40 & 400 &  \\
        CMU Pie & 1024 & 68 & 1564 &  \\
        \hline
        \multirow{2}{*} {Human Activity} & \multirow{2}{*} {561} & \multirow{2}{*} {6} & \multirow{2}{*} {5744} & Accelerometer \\
        & & & & Sensor\\
        \hline
        Isolet & 617 & 26 & 7797 & Speech\\
        \hline
        Mice Protein & 77 & 8 & 975 & {Biology}\\
        \hline
        10X PBMC & 16653 & 8 & 4271 &  \\
        Mouse ES Cell & 24175 & 4 & 2717 &  RNA-Seq \\
        Mouse Bladder Cell & 20670 & 16 & 2746 &  \\
        Worm Neuron Cell & 13488 & 10 & 4186 & \\
		\hline
	\end{tabular}	
	\caption{Descriptions of the data sets used for bench-marking experiments.}
	\label{table:dataDescription}
\end{table}
The various application domains make the set diverse; for example, some data sets (MNIST, Human Activity, Mice Protein, etc.) have more samples than features, whereas others have more features than samples (10X PBMC, Mouse ES Cells, Mouse Bladder Cells, etc), making the benchmarking experiments more challenging. The image data sets, e.g, ORL, COIl20, Yale Faces, CMU Pie etc, contain complex geometric, appearance, and reflectance characteristics and have been widely used in literature to validate the supervised DR models \cite{wang2023worst,ghosh2024linear}. Using these datasets we ran four suites of benchmarks and the specific experimental protocols are provided below.\\
\textbf{Pattern prediction at low-dimensional space:} This experiment compares our proposed method with Fisher LDA \cite{duda2006pattern}. The solution of Fisher LDA has the form $S_B w  =  \lambda S_W w$ ($S_B$ and $S_W$ are the between-class and within-class scatter matrices) which can be solved using a generalized eigenvalue solver. We used Matlab's GSVD command to get a solution. One can also obtain a solution using eigen decomposition of the matrix ${S_W}^{-1}S_B$ where the assumption is $S_w$ is invertible. We have used both solvers for LDA and compared them with our proposed ConvexLDA. The goal of this experiment is to compare each model's low-dimensional ($p \in \{2,3\}$) embedding on USPS\footnote{USPS data link: \url{http://www.cad.zju.edu.cn/home/dengcai/Data/FaceData.html} },MNIST\footnote{We took the MNIST data from the PyTorch package.}, Human Activity, Mice Protein, and ISOLET data sets\footnote{Mice Protein, Isolet, Human Activity are available at UCI Machine Learning Repository.}. We split each data set into a ratio of 80:20 of training and test partition, except for MNIST, which has a separate test set. After fitting all models in the training set, we projected the training and test samples in low-dimensional space and then calculated the $5-$NN test accuracy. We repeat the process 20 times and report the average accuracy with standard deviation. \\
\textbf{Prediction rate as a function of embedding dimension:}
Apart from comparing models in low-dimensional space, we carried out an experiment to show how the prediction rate varies over different embedding dimensions. We used four image data sets, Olivetti, ORL, YaleFace\footnote{YaleFace, ORL link: \url{http://www.cad.zju.edu.cn/home/dengcai/Data/FaceData.html}}, and CMU PIE \footnote{PIE data link: \url{https://github.com/somayjain/FaceRecognition/blob/master/Code/CMU-PIE_dataset/CMUPIEData.mat}}, and split them into a $80:20$ ratio of train and test. Each model is fitted on the training partition, and the $5-$NN classification is calculated using different embedding dimensions. The process is repeated 20 times, and the average accuracies are plotted for comparison. To run the experiment faster for LDA, we reduce the data dimensionality using Singular Value Decomposition.\\
\textbf{Comparison with SOTA LDA-variant:}
In the third suite of benchmarking experiment, we systematically compared ConvexLDA with several class discriminant methods, including Pairwise Accuracy Criterion (aPAC)\cite{loog2001multiclass}, Harmonic Mean for Subspace Selection (HMSS) \cite{bian2008harmonic}, Geometric Mean for Subspace Selection (GMSS) \cite{tao2008geometric}, Max-Min Distance Analysis (MMDA) \cite{bian2010max}, Weighted Harmonic Mean Trace-Ratio Discriminant Analysis (STRDA) \cite{li2017beyond}, Worst-case Linear Discriminant Analysis (WLDA) \cite{zhang2010worst},  Semi-definite Worst Case Linear Discriminant Analysis (SD-WLDA) \cite{li2015worst}, Heteroscedastic Max--Min Distance Analysis (HMMDA) \cite{su2018heteroscedastic}, Dimension Reduction via maximizing pairwise separation probability (DR-MSP) \cite{yang2019nonparametric}, Wasserstein Discriminant Analysis (WDA-eig) \cite{liu2020ratio}, Worst-case Discriminative Dimension Reduction (WDDR) \cite{wang2023worst}. Some of these methods are particularly designed to mitigate worst-case problem of Fisher LDA. We evaluated these models across eight publicly available datasets, encompassing four real-world RNA-Seq datasets\footnote{RNA-Seq data link: \url {https://github.com/ttgump/scDeepCluster/tree/master/scRNA-seq\%20data}} that capture the intricacies of complex biological processes \cite{kolodziejczyk2015technology}. Following the experimental protocol of Wang et al. \cite{wang2023worst}, we partitioned each data set into a $50:50$ ratio of training and test. First we ran a five-fold cross-validation on the training partition to determine the optimal embedding dimension and $\lambda$ for ConvexLDA. After that we fit the model on the training partition and, projected the training and test samples on the optimal embedding dimension. At last, we predicted the class label of the test samples using a $1-$NN classifier and reported the average with standard deviation over twenty runs. As done in \cite{wang2023worst}, PCA is applied to reduce the dimension of each data set while keeping 98\% of total variance.

\subsection{Results}
This section presents the outcomes of our benchmarking experiments, accompanied by a comprehensive and enriched discussion.\\
\textbf{Comparison between LDA vs Convex LDA:}
We start with low-dimensional classification results as shown in Table \ref{table:ConvexLDA_LDA_comparison}, containing datasets which have more samples than the number of features.
\begin{table}[!h]
\small
  \hspace{-1.25cm}
	\begin{tabular} {|c|c|c|c|c|c|}
 	\hline
  
	\multicolumn{6}{|c|} {Classification on embedding dimension = 2}\\
  \hline
  \multirow{1}{*}{Models} & \multicolumn{1}{c|} {USPS} & \multicolumn{1}{c|} {MNIST} & \multicolumn{1}{c|} {Activity} & \multicolumn{1}{c|} {Mice Protein} & \multicolumn{1}{c|} {ISOLET} \\
		
		\hline
		GSVD LDA & 54.92 $\pm$ 1.30 & 45.47 $\pm$ 0.75 & 63.88 $\pm$ 1.40 & 51.11 $\pm$ 3.96  &  32.86 $\pm$ 1.33 \\
		\hline
		Eig LDA & 55.89 $\pm$ 1.76 & 47.16 $\pm$ 0.82 & 51.39 $\pm$ 1.97 & 55.30 $\pm$ 6.77 &  32.66 $\pm$ 1.20\\
		\hline
        ConvexLDA & \textbf{61.08} $\pm$ \textbf{0.93} & \textbf{53.99} $\pm$ \textbf{2.22} & \textbf{64.00} $\pm$ \textbf{3.25} & \textbf{84.05} $\pm$ \textbf{2.80} & \textbf{49.17} $\pm$ \textbf{1.42} \\

		\hline \hline
    \multicolumn{6}{|c|} {Classification on embedding dimension = 3}\\
        \hline
    	GSVD LDA & 70.68 $\pm$ 0.95 & $65.10 \pm 0.75$ & $69.82 \pm 1.21$ &  $68.33 \pm 3.23$ & $50.08 \pm 1.91$ \\
		\hline
		Eig LDA & 70.32 $\pm$ 1.48 & $62.04 \pm 2.27$ & $63.01 \pm 3.18$ & $64.51 \pm 3.27$ & $50.69 \pm 2.71$ \\
		\hline
        ConvexLDA & \textbf{77.58} $\pm$ \textbf{1.06} & \textbf{72.38} $\pm$ \textbf{0.28} & \textbf{74.72} $\pm$ \textbf{1.62} & \textbf{92.57} $\pm$ \textbf{2.37} & \textbf{66.62} $\pm$ \textbf{1.59} \\

		\hline
	\end{tabular}
	\caption{$5$-NN classification accuracies ($\%$) on 2 and 3-dimensional space using two different solvers of LDA and ConvexLDA. Each dataset is divided into training and test by $80-20 \%$ ratio, followed by fitting all models on the training partition. At last we took the low-dimensional training and test samples to compute the accuracy. The best results are highlighted in bold.}	

	\label{table:ConvexLDA_LDA_comparison}
\end{table}
In all the cases, prediction rate improves significantly in three-dimensional embedding, which is not surprising. The pattern recognition accuracy is the lowest in ISOLET data across all the models, although ConvexLDA's performance is significantly better than LDA. ISOLST has $26$ classes, which are the spoken letters of the English alphabet. Some of these letters, e.g., 'B', 'D', and 'T', sound similar. Apart from it, there is a variability in pronunciation among person-to-person which added one more layer of complexity in predicting these digits in low-dimension. Among the two approaches of LDA, GSVD solver performed better in six out of ten cases. In some cases, e.g., Activity in 2D and 3D, Mice Protein and MNIST in 3D, its performance is significantly better which suggest GSVD LDA may be a better option in datasets which have more samples than features while performing low-dimensional classification. Compared to both the solvers of LDA, our proposed method consistently produce better recognition rate across all the data sets. In all the cases, ConvexLDA performed significantly better than LDA, which suggest our approach is better at keeping similar objects tightly clustered and separating classes from each other with out collapsing them in low-dimension, i.e., ConvexLDA overcomes the \textit{class collapse} issue of Fisher LDA. 
\begin{figure}[!h]
	\includegraphics[width=12.0cm,height=8.75cm]{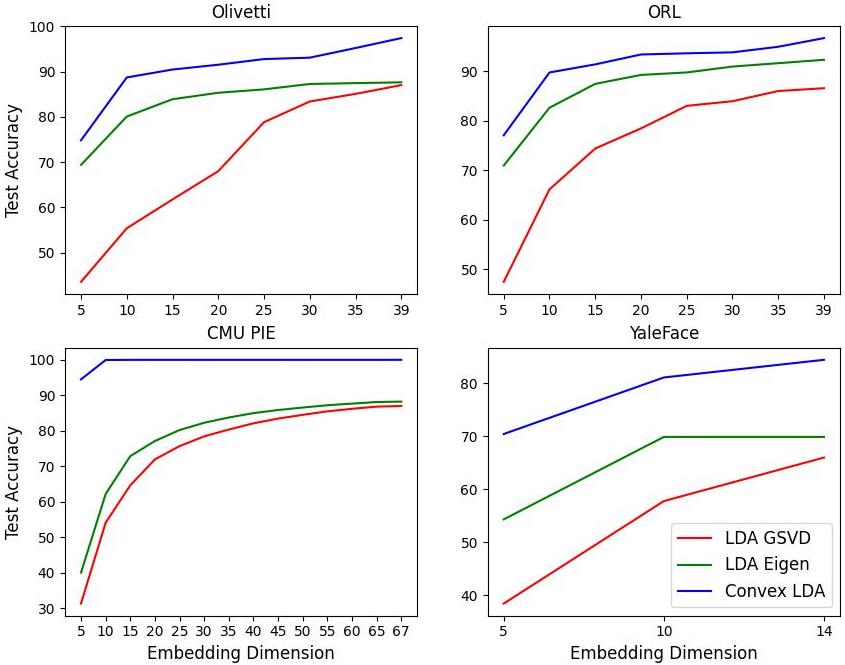}
	\caption{Test sample prediction accuracy as a function of embedding dimension on Olivetti, ORL, CMU PIE, and Yale Face images. In each case the embedding dimension is varied from 5 to (\# classes -1) which is the maximum projection dimension of LDA. Each data set is partitioned $80-20$ ratio of training and test and the out of sample accuracy is averaged over 20 runs using a $5$-NN classifier.}
	\label{fig:Accuracy_vs_embeddingDim}
\end{figure}
We further compared the recognition rate of test samples as a function of the embedding dimension between the range 5 to the maximum projection dimension of LDA which is (\# classes -1) to make the comparison fair. Note, ConvexLDA doesn't have such a restriction on maximum projection dimension. To this end we chose image datasets, e.g., Olivetti, ORL, CMU PIE, and Yale Face, which are used extensively in the literature to compare supervised linear models \cite{ghosh2024linear,wang2023worst}. Apart from containing numerous classes, ranging from 15 to 68, these human face datasets are collected under different illumination, poses, facial expression (e.g., sad, angry, happy, etc), etc, which make the recognition task challenging. Figure \ref{fig:Accuracy_vs_embeddingDim} shows the result. Out of sample recognition accuracy increases across all the models with the increase of embedding dimension. Among the two solvers of LDA, solution using eigen-decomposition outperformed the GSVD solver in all cases. We think the orthogonal constraint among the solution vectors perhaps the reason of improved the generalization performance of eigen-decomposition-based solver. The performance of Convex LDA is the best in all the four datasets across all embedding dimensions. Particularly in CMU Pie data, our proposed method achieved a zero error rate staring from dimension 15, leaving behind LDA Eigen by a margin of more than $25\%$. The accuracy curve of other three datasets suggests that the performance of Convex LDA has not saturated yet, and adding more dimension will improve the prediction rate further.\\
\textbf{Comparison between ConvexLDA vs Several SOTA DR Methods:}
Now, we present the result comparing our approach to several state-of-the-art linear DR methods which are proposed to mitigate the class-collapse issue of LDA. The comparison is done in two experiments. The first experiment includes data sets where the number of samples is higher than the data dimension (except for ORL). All models were run using optimal embedding dimension. In contrast, the second experiment was conducted on three fixed embedding dimension on four high-dimensional RNA-seq data. Table \ref{table:ConvexLDA_SOTA_comparison1} presents the result of the first experiment where all the models achieved relatively good prediction rate. Note that, most of the methods did better than the Fisher LDA, which is not surprising as they are innovated to mitigate the drawbacks of LDA. In COIL20, the differences in recognition rate among the models are slim and all of them exceed the $95\%$ threshold. This dataset contains images of several toys, (e.g., robot, train, plane), scissors, light bulb, wrench, etc, which are different looking 
\begin{table*}[ht!]
        \hspace{-2.5 cm}
	\begin{tabular} {|c|c|c|c|c|c|c|}
 	\hline
	\multirow{2}{*}{Models} & \multicolumn{6}{|c|} {Data Sets }\\
  \cline{2-7}
   & \multicolumn{1}{c|} {ISOLET} & \multicolumn{1}{c|} {USPS} & \multicolumn{1}{c|} {MNIST} & \multicolumn{1}{c|} {MSRA25} & \multicolumn{1}{c|} {COL20} & \multicolumn{1}{c|} {ORL} \\
	\hline
        LDA & $85.14 \pm 1.24$ & $83.92 \pm 0.79$ & $82.60 \pm 0.46$ & $87.54 \pm 0.74$ & $96.16 \pm 1.27$ & $90.36 \pm 1.64$ \\
        \hline
        aPAC & $87.66 \pm 1.61$ & $85.17 \pm 1.47$ & $84.61 \pm 0.42$ & $90.55 \pm 0.87$ & $96.23 \pm 1.01$ & $88.81 \pm 2.91$ \\
        \hline
        GMSS & $88.43 \pm 1.84$ & $85.27 \pm 1.36$ & $84.00 \pm 0.74$ & $90.64 \pm 0.80$ & $96.49 \pm 1.06$ & $88.93 \pm 3.79$ \\
        \hline
        HMSS & $88.33 \pm 1.37$ & $84.90 \pm 0.97$ & $84.22 \pm 0.67$ & $90.70 \pm 0.77$ & $96.34 \pm 1.00$ & $88.81 \pm 3.98$ \\
        \hline
        MMDAL & $88.51 \pm 1.51$ & $85.06 \pm 1.01$ & $84.01 \pm 0.52$ & $90.71 \pm 0.70$ & $96.47 \pm 1.06$ & $88.93 \pm 3.98$ \\
        \hline
        MMDAG & $88.38 \pm 1.53$ & $85.03 \pm 1.34$ & $84.01 \pm 0.52$ & $90.71 \pm 0.70$ & $96.37 \pm 1.05$ & $88.93 \pm 3.98$ \\
        \hline
        STRDA & $85.91 \pm 2.37$ & $87.93 \pm 1.15 $ & $86.85 \pm 0.41$ & $90.54 \pm 0.63$ & $95.65\pm± 0.88$ & $91.90 \pm 1.49$ \\
	\hline
        WLDA & $88.51 \pm 1.51$ & $85.06 \pm 1.07$ & $84.69 \pm 0.74$ & $90.39 \pm 0.78$ & $95.97 \pm 1.07$ & $88.45 \pm 3.08$ \\
        \hline    
        SD-WLDA & $89.98 \pm 1.34$ & $87.18 \pm 1.12$ & $86.33 \pm 0.62$ & $91.03 \pm 0.21$ & $96.85 \pm 1.09$ & $90.36 \pm 2.39$ \\
        \hline
        HMMDA & $89.26 \pm 1.12$ & $86.34 \pm 1.27 $ & $ 85.49 \pm 0.23$ & $91.01 \pm 0.85$ & $96.89 \pm 2.01$ & $90.15 \pm 3.22$ \\
	\hline
        DR-MSP & $88.43 \pm 1.98$ & $86.68 \pm 0.99$ & $86.01 \pm 0.47$ & $86.68 \pm 0.99$ & $96.12 \pm 1.09$ & $89.87 \pm 2.01$ \\
	\hline
        WDA-eig & $89.03 \pm 1.01$ & $86.99 \pm 1.01$ & $86.28 \pm 0.21$ & $86.99 \pm 1.01$ & $96.21 \pm 1.43$ & $90.02 \pm 2.23$ \\
	\hline
        WDDR$^{1}$ & $92.14 \pm 0.93$ & $88.87 \pm 1.15$ & $87.56 \pm 0.47$ & $91.66 \pm 0.80$ & $97.32 \pm 0.34$ & $92.14 \pm 1.56$ \\
        \hline
        WDDR$^{2}$ & $91.57 \pm 0.84$ & $89.02 \pm 1.34$ & $87.01 \pm 0.39 $ & $91.58 \pm 0.73$ & $97.08 \pm 0.29$ & $93.19 \pm 1.21$ \\
	\hline
	ConvexLDA & $\textbf{93.35} \pm \textbf{0.36}$ & $\textbf {94.91} \pm \textbf {0.42}$ & $\textbf{93.93} \pm \textbf{0.22}$ & $\textbf{99.89} \pm \textbf{0.11}$ & $\textbf{98.44} \pm \textbf{0.74}$ & $\textbf{94.08} \pm \textbf{1.89}$ \\
    \hline
	\end{tabular}
	\caption{ Comparison of $1$-NN classification accuracies ($\%$) on six publicly available datasets using fifteen supervised DR methods using their optimal embedding dimension. In each case the data is divided into equal size of training and test sets. The models are trained on the training partition and then $1$-NN accuracy is computed using the test set to report the average prediction rate over twenty iterations. The best results are highlighted in bold. The results other than ConvexLDA are being reported from \cite{wang2023worst}. }	

	\label{table:ConvexLDA_SOTA_comparison1}
\end{table*}
objects and as a result the features are discriminative enough for high recognition rate across the models. It's noteworthy, that in ORL, LDA performed bettter than aPAC, GMSS, HMSS, MMDAL, MMDAG, WLDA, and DR-SMP. In all the cases, ConvexLDA reached the highest prediction accuracy, leaving behind recently proposed WDDRs \cite{wang2023worst}. In some cases, e.g., USPS, MNIST, and MSRA25, ConvexLDA outperformed WDDRs by a margin between $6\% - 9\%$. Clearly, the convex nature of our proposed method found a better global optimum compared to WDDRs which is designed to separate the worst class-pairs. 
In contrast, our method maximizes scatter across all class pairs by adjusting the hyperparameter $\lambda$. Unlike WDDR, which prioritizes separating the worst-class pairs, our approach applies a global separation force across all classes, leading to more balanced class dispersion. 
\begin{table*}[h!]
    \hspace{-2.5cm}
	\begin{tabular} {|c|c|c|c|c|c|c|c|c|c|c|c|c|}
 	\hline
	\multirow{3}{*}{Models} & \multicolumn{12}{|c|} {Data Sets }\\
  \cline{2-13}
   & \multicolumn{3}{c|} {10X PBMC} & \multicolumn{3}{c|} {Mouse ES Cell} & \multicolumn{3}{c|} {Mouse Bladder Cell} & \multicolumn{3}{c|} {Mouse Neuron Cell} \\
		\cline{2-13}
   & \multicolumn{3}{c|} {Embedding Dimension} & \multicolumn{3}{c|} {Embedding Dimension} & \multicolumn{3}{c|} {Embedding Dimension} & \multicolumn{3}{c|} {Embedding Dimension} \\
		\cline{2-13}
    & 30 & 70 & 100 & 30 & 70 & 100 & 30 & 70 & 100 & 30 & 70 & 100\\
\hline

        LDA & 45.31 & 46.59 & 46.71 & 55.89 & 56.10 & 55.76 & 42.13 & 44.32 & 44.10 & 46.23 & 46.49 & 47.01 \\
        \hline
        aPAC & 48.49 & 49.13 & 48.93 & 57.67 & 57.19 & 56.49 & 42.36 & 44.21 & 45.01 & 42.12 & 42.36 & 41.94 \\
        \hline
        GMSS & 52.63 & 52.92 & 52.29 & 58.82 & 59.27 & 58.33 & 43.11 & 44.68 & 44.29 & 43.59 & 42.34 & 43.29 \\
        \hline
        HMSS & 52.02 & 52.38 & 53.03 & 60.25 & 60.24 & 61.33 & 44.17 & 45.32 & 46.28 & 41.30 & 41.38 & 42.02 \\
        \hline
        MMDAL & 55.78 & 55.90 & 55.28 & 65.83 & 66.02 & 65.39 & 47.33 & 47.49 & 48.27 & 48.31 & 49.01 & 50.10 \\
	\hline
        MMDAG & 55.43 & 56.09 & 56.94 & 65.12 & 66.25 & 65.93 & 47.55 & 47.32 & 48.34 & 48.21 & 49.11 & 49.96 \\
	\hline
        STRDA & 56.21 & 57.90 & 57.23 & 67.44 & 67.39 & 68.02 & 50.11 & 50.24 & 49.66 & 50.38 & 51.02 & 51.22 \\
	\hline
        WLDA & 55.92 & 56.31 & 56.92 & 67.31 & 68.03 & 67.38 & 50.03 & 50.16 & 50.25 & 51.32 & 51.05 & 50.95 \\
        \hline
        SD-WLDA & 58.61 & 58.19 & 59.10 & 68.36 & 67.27 & 68.03 & 50.39 & 51.34 & 51.21 & 52.44 & 52.01 & 52.13 \\
        \hline
        HMMDA & 55.81 & 55.23 & 56.03 & 65.22 & 65.38 & 65.31 & 49.93 & 49.30 & 50.01 & 51.45 & 51.04 & 52.00 \\
	\hline
        DR-MSP & 57.22 & 56.98 & 57.28 & 68.01 & 68.28 & 67.93 & 51.44 & 51.39 & 52.02 & 52.04 & 52.82 & 52.09 \\
	\hline
        WDA-eig & 50.37 & 51.32 & 52.04 & 57.01 & 56.43 & 56.19 & 46.92 & 47.03 & 47.12 & 49.51 & 50.11 & 50.38 \\
	\hline
        WDDR$^{1}$ & 61.05 & 62.18 & 61.98 & 70.13 & 70.02 & 69.69 & 52.83 & 53.12 & 53.45 & 54.38 & 54.49 & 53.28 \\
        \hline
        WDDR$^{2}$ & 60.21 & 62.02 & 62.18 & 71.62 & 71.33 & 70.43 & 52.19 & 53.01 & 53.26 & \textbf{55.01} & 55.39 & 55.03 \\
	\hline
        \multirow{2}{*}{ConvexLDA} & \textbf{83.40} & \textbf{85.35} & \textbf{85.74} & \textbf{90.69} & \textbf{93.60} & \textbf{93.35} & \textbf{77.55} & \textbf{81.17} & \textbf{82.36} & 44.15 & \textbf{56.55} & \textbf{61.20} \\
        
         & $\pm 0.76$ & $\pm 0.77$ & $\pm 0.79$ & $\pm 1.69$ & $\pm 1.09$ & $\pm 0.82$ & $\pm 1.32$ & $\pm 1.22$ & $\pm 0.80$ & $\pm 1.23$ & $\pm 1.97$ & $\pm 1.62$ \\
  
		\hline
	\end{tabular}
	\caption{ Comparison of $1$-NN prediction rates (in $\%$) on test samples on four high-dimensional biological datasets using fifteen supervised DR methods using their optimal embedding dimension. Each model is fit on training set by taking $50\%$ of total samples randomly. After that test set accuracy is computed on embedding space of each model. The best results are highlighted in bold. The results other than ConvexLDA are being reported from \cite{wang2023worst}, which doesn't report the standard deviation. }	

	\label{table:ConvexLDA_SOTA_biodata}
\end{table*}
We hypothesize that excessive emphasis on worst-class pair separation, as in WDDR, may improve their separability but inadvertently reduce the overall scatter of other class pairs, potentially increasing the classification error rate.

Next, we evaluate the results on four high-dimensional real-world biological datasets, as presented in Table \ref{table:ConvexLDA_SOTA_biodata}. Compared to the previous experiment, the overall classification accuracies are lower, reflecting the inherent challenges of single-cell RNA-seq data, such as sparsity, high dimensionality, and noise in data collection. The relatively poor performance of LDA across the first three datasets suggests that it may not be well-suited for handling these challenges. It has been reported that in high-dimensional settings, the within-class scatter matrix $S_w$ often becomes singular \cite{witten2011penalized}, leading to performance degradation.

ConvexLDA achieved the highest prediction rate in eleven out of twelve cases, demonstrating its broad applicability beyond image data. Models designed to optimize worst-class pair separation, such as WLDA, DR-MSP, and WDDRs, performed better than LDA but still fell significantly short compared to ConvexLDA. Notably, in the PBMC, ES, and Bladder cell datasets, WDDRs—the second-best-performing approach—lagged behind ConvexLDA by a margin of 20-25\%. This further supports the effectiveness of applying a global separation force across all class pairs, rather than exclusively maximizing the distance of the worst-class pairs. These results reinforce the strength of ConvexLDA in handling overlapping and heterogeneous classes in high-dimensional biological data.


\section{Conclusion and Future Work}
\label{conc_future_work}

In this article, we introduced an alternative formulation of LDA that avoids directly optimizing the ratio of between-class to within-class scatter. Instead, we propose a weighted multi-objective cost function, where the first objective maps each sample to its class centroid in the low-dimensional space, while the second maximizes the scatter of these centroids. The weighting parameter, $\lambda$, serves as a control mechanism for class separation, with higher values increasing inter-class separation and lower values maintaining closer proximity. 
An in-depth analysis using synthetic data, evaluated through metrics such as Hausdorff distance, class scatter (CS), and mean set distance (MSD), provides strong empirical evidence supporting our approach. Furthermore, benchmarking experiments on real-world datasets further validate the effectiveness of the proposed method.

Our analysis demonstrates that this approach effectively mitigates worst-case class collapse, a common limitation of LDA. While alternative methods attempt to maximize worst-class pair distances, they often risk bringing other class pairs closer. In contrast, ConvexLDA applies a global separation force, ensuring uniform class separation. Classification results across multiple datasets in two, three, and optimally determined embedding dimensions validate our claims. ConvexLDA not only outperforms conventional LDA but also surpasses state-of-the-art models such as WLDA, STRDA, SD-WLDA, DR-MSP, WDA-eig, and WDDR, which are specifically designed for worst-class separation. Moreover, in experiments on four high-dimensional biological datasets, ConvexLDA significantly outperformed LDA and other models, demonstrating its effectiveness in the high-dimension, low-sample-size (HDLSS) regime.

Beyond achieving a higher prediction rate, the convex nature of our method guarantees global optimality. Its gradient-based optimization converges efficiently in high-dimensional settings, whereas LDA relies on the eigen-decomposition of a $d \times d$ scatter matrix—a process that becomes prohibitively slow as $d$ increases. Many worst-class separation methods, such as WDDR, MMDA, and WLDA, rely on semi-definite programming (SDP), which is computationally expensive. Given the inefficiency of SDP solvers \cite{flicker2020comparison}, we anticipate that ConvexLDA will provide significant computational advantages over these approaches.

Despite its advantages, ConvexLDA has certain limitations. As a linear transformation model, it does not capture nonlinear relationships in data. In cases where nonlinearity is essential, the objective function could be optimized within a neural network framework. Additionally, ConvexLDA relies on the $\ell_2$-norm, making it potentially sensitive to outliers. Future research could explore incorporating an $\ell_1$-norm in the first term to improve robustness against outliers.

Although the model demonstrates strong predictive performance on high-dimensional biological datasets, it is not inherently designed for biomarker selection. However, this capability could be integrated by introducing sparsity-promoting regularization, such as an $\ell_1$-norm penalty term. Exploring these extensions will be a promising avenue for future research to further enhance the applicability of ConvexLDA.

\bibliographystyle{plain}
\bibliography{ConvexLDA}

\end{document}